\documentclass{llncs}
\usepackage{times}
\usepackage{helvet}
\usepackage{courier}
\usepackage{adjustbox}
\usepackage{booktabs} 
\usepackage{amssymb}
\usepackage{float}
\begin{document}
	
	\title{Learning to Predict: A Fast Re-constructive Method to Generate Multimodal Embeddings}
	
	

\author{Guillem Collell\inst{1} \and Ted Zhang\inst{2} \and Marie-Francine Moens\inst{3}}
\institute{Computer Science Department, KU Leuven, Belgium\inst{1}\inst{2}\inst{3}\\
\email{gcollell@kuleuven.be, tedz.cs@gmail.com, sien.moens@cs.kuleuven.be} }

\maketitle   
	
	
\begin{abstract}
	Integrating visual and linguistic information into a single multimodal representation is an unsolved problem with wide-reaching applications to both natural language processing and computer vision. In this paper, we present a simple method to build multimodal representations by learning a language-to-vision mapping and using its output to build multimodal embeddings. In this sense, our method provides a cognitively plausible way of building representations, consistent with the inherently re-constructive and associative nature of human memory. Using seven benchmark concept similarity tests we show that the mapped vectors not only implicitly encode multimodal information, but also outperform strong unimodal baselines and state-of-the-art multimodal methods, thus exhibiting more ``human-like" judgments---particularly in zero-shot settings.
\end{abstract}

\section{Introduction}
\label{sect:intro}

Convolutional neural networks (CNN) and distributional-semantic models have provided breakthrough advances in representation learning in computer vision (CV) and natural language processing (NLP) respectively \cite{lecun2015deep}. Lately, a large body of research has shown that using rich, multimodal representations created from combining textual and visual features instead of unimodal representations (a.k.a. embeddings) can improve the performance of semantic tasks. In other words, a single multimodal representation that captures information from two modalities (vision and language) is semantically richer than those from a single modality or unimodal (either vision or language). Building multimodal representations has become a popular problem in NLP that has yielded a wide variety of methods \cite{lazaridou2015combining,kiela2014learning,silberer2014learning}. Additionally, the use of a mapping to bridge vision and language has also been explored, typically with the goal of zero-shot image classification \cite{lazaridou2014wampimuk,socher2013zero}.

\begin{figure}
	\begin{center}
		\includegraphics[scale=0.56]{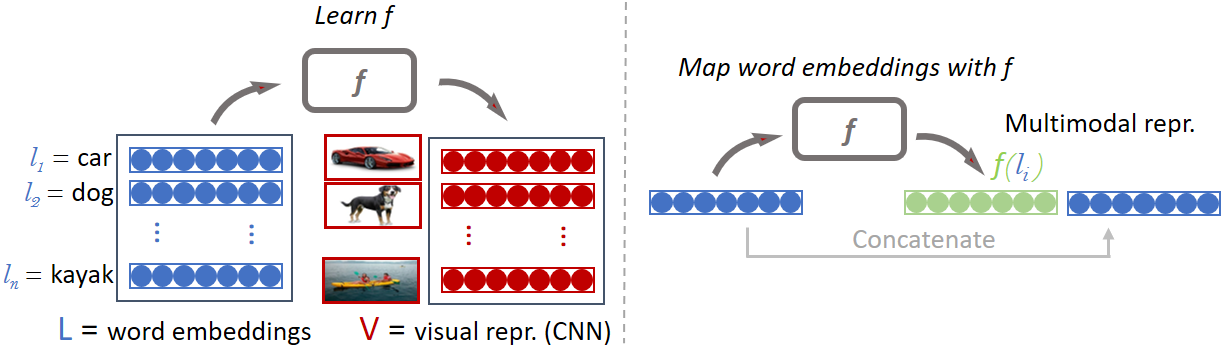}
	\end{center}
	\caption{Overview of our multimodal method.}
	\label{diagram}
\end{figure}

Here, we propose a cognitively plausible approach to concept representation that consists of: (1) learning a language-to-vision mapping; and (2) using the outputs of the mapping as multimodal representations---with the second step being the main novelty of our approach. By re-constructing visual knowledge from textual input, our method behaves similarly as human memory, namely in an associative \cite{anderson2014human,reijmers2007localization} and re-constructive manner \cite{vernon2014artificial,hawkins2007intelligence,loftus1981reconstructive}. Concretely, our method does not seek the perfect recall of visual representations but rather its re-construction and association with language. We leverage the intuitive fact that, by learning to predict, the mapping necessarily encodes information from both modalities---and in turn discards noise and irrelevant information from the visual vectors during the learning phase. Thus, given a word embedding as input, the mapped output is not purely a visual representation but rather a multimodal one.   

By using seven concept similarity benchmarks, we show that our representations not only are multimodal but they improve performance over strong unimodal baselines and state-of-the-art multimodal approaches---inclusive in a zero-shot setting. In turn, the fact that our evaluation tests are composed of human ratings of similarity supports our claim that our method provides more ``human-like" judgments. Further details and insight can be found at the extended version of the present paper \cite{collell2017imagined}. 

The rest of the paper is organized as follows. In the next section, we introduce related work. Next, we describe and provide insight on our method. Afterwards, we describe our experimental setup. Finally, we discuss our results, followed by conclusions.

\section{Related work and background}
\label{sect:related}

\subsection{Cognitive grounding}
\label{sect:cognitive}
A large body of research evidences that human memory is inherently re-constructive \cite{vernon2014artificial,hawkins2007intelligence,loftus1981reconstructive}. That is, memories are not ``static" exact copies of reality, but are rather re-constructed from their essential elements each time they are retrieved, triggered by either internal or external stimuli. Arguably, this mechanism is, in turn, what endows humans with the capacity to imagine themselves in yet-to-be experiences and to re-combine existing knowledge into new plans or structures of knowledge \cite{hawkins2007intelligence}. Moreover, the associative nature of human memory is also a widely accepted theory in experimental psychology \cite{anderson2014human} with identifiable neural correlates involved in both learning and retrieval processes \cite{reijmers2007localization}.

In this respect, our method employs a retrieval process analogous to that of humans, in which the retrieval of a visual output is triggered and mediated by a linguistic input (Fig. \ref{diagram}). Effectively, visual information is not only retrieved (i.e., mapped), but also associated to the textual information thanks to the learned cross-modal mapping---analogous to a mental model that associates semantic and visual components of concepts, acquired through lifelong experience. Since the retrieved (mapped) visual information is often insufficient to completely describe a concept, it is of interest to preserve the linguistic component. Thus, we consider the concatenation of the ``imagined" visual representations to the text representations as a comprehensive way of representing concepts.

\subsection{Multimodal representations}
\label{sect:multimodal}
It has been shown that visual and textual features capture complementary attributes \cite{collell2016is}, and the advantages of combining both modalities have been largely demonstrated in a number of linguistic tasks \cite{lazaridou2015combining,kiela2014learning,silberer2014learning}. Based on current literature, we suggest a classification of the existing strategies to build multimodal embeddings. Broadly, multimodal representations can be built by learning from raw input enriched with both modalities (\textbf{\textit{simultaneous}} learning), or by learning each modality separately and integrating them afterwards (\textbf{\textit{a posteriori}} combination).

\begin{enumerate}
	\item \textbf{\textit{A posteriori}} combination.
	\begin{itemize}
		\item \emph{Concatenation}. That is, the fusion of pre-learned visual and text features by concatenating them \cite{kiela2014learning}. Concatenation has been proven effective in concept similarity tasks \cite{bruni2014multimodal,kiela2014learning}, yet suffers from an obvious limitation: multimodal features can only be generated for those words that have images available.
		\item \emph{Autoencoders} form a more elaborated approach that do not suffer from the above problem. Encoders are fed with pre-learned visual and text features, and the hidden representations are then used as multimodal embeddings. This approach has shown to perform well in concept similarity tasks and categorization (i.e., grouping objects into categories such as ``fruit", ``furniture", etc.) \cite{silberer2014learning}.
		\item A \emph{mapping} between visual and text modalities (i.e., our method). The outputs of the mapping themselves are used to build multimodal representations.
	\end{itemize}
	\item \textbf{\textit{Simultaneous}} learning. Distributional semantic models are extended into the multimodal domain \cite{lazaridou2015combining,hill2014learning} by learning in a skip-gram manner from a corpus enriched with information from both modalities and using the learned parameters of the hidden layer as multimodal representations. Multimodal skip-gram methods have been proven effective in similarity tasks \cite{lazaridou2015combining,hill2014learning} and in zero-shot image labeling \cite{lazaridou2015combining}.
\end{enumerate}

With this taxonomy, the gap that our method fills becomes more clear, with it being aligned with a re-constructive and associative view of knowledge representation. Furthermore, in contrast to other multimodal approaches such as skip-gram methods \cite{lazaridou2015combining,hill2014learning}, our method directly learns from pre-trained embeddings instead of training from a large multimodal corpus, rendering it thus simpler and faster.

\subsection{Cross-modal mappings}
\label{sect:crossmodalmap}
Several studies have considered the use of mappings to bridge modalities. For instance, \cite{socher2013zero} and \cite{lazaridou2014wampimuk} use a linear vision-to-language projection in zero-shot image classification. Analogously, language-to-vision mappings have been considered, generally to generate missing perceptual information about abstract words \cite{hill2014learning,johns2012perceptual} and in zero-shot image retrieval \cite{lazaridou2015combining}. In contrast to our approach, the methods above do not aim to build multimodal representations to be used in natural language processing tasks.

\section{Proposed method}
In this section we describe the three main steps of our method (Fig. \ref{diagram}): (1) Obtain visual representations of concepts; (2) Build a mapping from the linguistic to the visual space; and (3) Generate multimodal representations.

\subsection{Obtaining visual representations}
\label{sect:visrep}

We employ raw, labeled images from ImageNet \cite{russakovsky2015imagenet} as the source of visual information, although alternatives such as the ESP game data set \cite{von2004labeling} can be considered. To extract visual features from each image, we use the forward pass of a pre-trained CNN model. The hidden representation of the last layer (before the softmax) is taken as a feature vector, as it contains higher level features. For each concept $w$, we \textit{average} the extracted visual features of individual images to build a single visual representation $\overrightarrow{v_w}$.

\subsection{Learning to map language to vision}
\label{sect:mapping}

Let $\mathcal{L} \subset \mathbb{R}^{d_l}$ be the linguistic space and $\mathcal{V} \subset \mathbb{R}^{d_v}$ the visual space of representations, where $d_l$ and $d_v$ are their respective dimensionalities. Let $\overrightarrow{l_w} \in \mathcal{L}$ and $\overrightarrow{v_w} \in \mathcal{V}$ denote the text and visual representations for the concept $w$ respectively. Our goal is thus to learn a mapping (regression) $f:\mathcal{L}\rightarrow \mathcal{V}$. The set of $N$ visual representations along with their corresponding text representations compose the training data $\{ (\overrightarrow{l_i},\overrightarrow{v_i})\}^{N}_{i=1}$ used to learn $f$. In this work, we consider two different mappings $f$.

\textbf{(1) Linear:} A simple perceptron composed of a $d_l$-dimensional input layer and a linear output layer with $d_v$ units.

\textbf{(2) Neural network:} A network composed of a $d_l$-unit input layer, a single hidden layer of $d_h$ Tanh units and a linear output layer of $d_v$ units.

For both mappings, a mean squared error (MSE) loss function is employed: $Loss(y,\hat{y}) = \frac{1}{2} ||\hat{y} - y||^2_2 $, where $y$ is the actual output and $\hat{y}$ the model prediction. 

\subsection{Generating multimodal representations}
\label{sect:mapped}

Finally, the mapped representation $\overrightarrow{m_w}$ of each concept $w$ is calculated as the image $f(\overrightarrow{l_w})$ of its linguistic embedding $\overrightarrow{l_w}$. For instance, $\overrightarrow{m_{dog}} = f(\overrightarrow{l_{dog}})$. We henceforth refer to the mapped representations as \textit{MAP}$_{f}$, where $f$ indicates the mapping function employed ($lin$ = linear, $NN$ = neural network). As argued below, the mapped representations are effectively multimodal. However, since $f(\overrightarrow{l_w})$ formally belongs to the visual domain, we also consider the concatenation of the $\ell_2$-normalized mapped representations $f(\overrightarrow{l_w})$ with the textual representations $\overrightarrow{l_w}$, namely $\overrightarrow{l_w}\oplus f(\overrightarrow{l_w})$, where $\oplus$ denotes the concatenation operator. We denote these concatenated representations as \textit{MAP-C}$_{f}$.

Since the outputs of a text-to-vision mapping are strictly speaking, ``visual predictions", it might not seem readily obvious that they are also grounded with textual knowledge. To gain insight, it is instructive to refer to the training phase where the parameters $\theta$ of $f$ are learned as a function of the training data $\{(\overrightarrow{l_i},\overrightarrow{v_i})\}^{N}_{i=1}$. E.g., in gradient descent, $\theta$ is updated according to: $\theta \leftarrow \theta - \eta \frac{\partial}{\partial \theta } Loss( \theta ; \{(\overrightarrow{l_i},\overrightarrow{v_i})\}^{N}_{i=1}).$ Hence, the parameters $\theta$ of $f$ are effectively a function of the training data points $\{(\overrightarrow{l_i},\overrightarrow{v_i})\}^{N}_{i=1}$ and it is therefore expected that the outputs $f(\overrightarrow{l_w})$ are grounded with properties of the input data $\{\overrightarrow{l_i}\}^{N}_{i=1}$. It can be additionally noted that the output of the mapping $f(\overrightarrow{l_w})$ is a (continuous) transformation of the input vector $\overrightarrow{l_w}$. Thus, unless the mapping is completely uninformative (e.g., constant or random), the input vector $\overrightarrow{l_w}$ is still ``present"---yet transformed. Thus, the output of the mapping necessarily contains information from both modalities, vision and language, which is essentially the core idea of our method. Further insight is provided at the extended version of the article \cite{collell2017imagined}.

\section{Experimental setup} 
\subsection{Word embeddings}
We use 300-dimensional GloVe\footnote{\tt http://nlp.stanford.edu/projects/glove} vectors \cite{pennington2014glove} pre-trained on the Common Crawl corpus consisting of 840B tokens and a 2.2M words vocabulary. 

\subsection{Visual data and features}
We use ImageNet \cite{russakovsky2015imagenet} as our source of labeled images. ImageNet covers 21,841 WordNet synsets (or meanings) \cite{fellbaum1998wordnet} and has 14,197,122 images. We only keep synsets with more than 50 images, and an upper bound of 500 images per synset is used to reduce computation time. With this selection, we cover 9,251 unique words.

To extract visual features from each image, we use a pre-trained VGG-m-128 CNN \cite{Chatfield14} implemented with the Matlab MatConvNet toolkit \cite{vedaldi15matconvnet}. We take the 128-dimensional activation of the last layer (before the softmax) as our visual features.

\subsection{Evaluation sets}
We test the methods in seven benchmark tests, covering three tasks: \textbf{(i) General relatedness}: \textit{MEN} \cite{bruni2014multimodal} and \textit{Wordsim353-rel} \cite{agirre2009study}; \textbf{(ii) Semantic or taxonomic similarity}: \textit{SemSim} \cite{silberer2014learning}, \textit{Simlex999} \cite{hill2015simlex}, \textit{Wordsim353-sim} \cite{agirre2009study} and \textit{SimVerb-3500} \cite{gerz2016simverb}; \textbf{(iii) Visual similarity}: \textit{VisSim} \cite{silberer2014learning} which contains the same word pairs as \textit{SemSim}, rated for visual instead of semantic similarity. All tests contain word pairs along with their human similarity rating. The tests \textit{Wordsim353-sim} and \textit{Wordsim353-rel} are the similarity and relatedness subsets of \textit{Wordsim353} \cite{finkelstein2001placing} proposed by \cite{agirre2009study} who noted that the distinction between similarity (e.g., ``tiger" is similar to ``cat") and relatedness (e.g., ``stock" is related to ``market") yields different results. Hence, for being redundant with its subsets, we do not count the whole \textit{Wordsim353} as an extra test set.

A large part of words in our tests do not have a visual representation $\overrightarrow{v_w}$ available, i.e., they are not present in our training data. We refer to these words as zero-shot (ZS).

\subsection{Evaluation metric and prediction}

We use Spearman correlation $\rho$ between model predictions and human similarity ratings as evaluation metric. The prediction of similarity between two concept representations, $\overrightarrow{u_1}$ and $\overrightarrow{u_2}$, is computed by their cosine similarity: $\cos(\overrightarrow{u_1},\overrightarrow{u_2}) = \frac{\overrightarrow{u_1} \cdot \overrightarrow{u_2}}{\| \overrightarrow{u_1} \|\cdot\| \overrightarrow{u_2} \|}$.

\subsection{Model settings}

Both, neural network and linear models are learned by stochastic gradient descent and nine parameter combinations are tested (learning\_rate = [0.1, 0.01, 0.005] and dropout\_rate = [0.5, 0.25, 0.1]). We find that the models are not very sensitive to parameter variations and all of them perform reasonably well. We report a linear model with learning rate of 0.1 and dropout rate of 0.1. For the neural network we use 300 hidden units, dropout rate of 0.25 and learning of 0.1. All mappings are implemented with the scikit-learn toolkit \cite{scikit-learn} in Python 2.7.

\section{Results and discussion} 
\label{sect:results}

In the following we summarize our main findings. For clarity, we refer to the concatenation of \textit{CNN}$_{avg}$ and GloVe as \textit{CONC}. 

\textbf{\textit{Overall}}, a post-hoc Nemenyi test including all disjoint regions (ZS and VIS) shows that both \textit{MAP-C} methods (\textit{lin} and \textit{NN}) perform significantly better than GloVe (p $\approx$ 0.03) and than \textit{CNN}$_{avg}$ (p $\approx$ 0.06). Hence, our multimodal representations \textit{MAP-C} clearly accomplish one of their foremost goals, namely to improve the unimodal representations of GloVe and \textit{CNN}$_{avg}$. 

Clearly, the consistent improvement of \textit{MAP}$_{lin}$ and \textit{MAP}$_{NN}$ over \textit{CNN}$_{avg}$ in all seven test sets supports our claim that the \textit{imagined} visual representations are more than purely visual representations and contain multimodal information---as argued in subsection \ref{sect:mapped}. Moreover, the \textit{MAP-C} method generally performs better than the\textit{MAP} vectors alone, implying that even though the \textit{MAP} vectors are indeed multimodal, they are still predominantly visual and their concatenation with textual representations helps.

Using the \textit{\textbf{concreteness}} ratings of \cite{brysbaert2014concreteness} in a 1-5 scale (with 5 being the most concrete and 1 the most abstract) we find that the average concreteness is larger than 4.4 in all VIS regions, while it is lower than 3.3 in all ZS regions except in \textit{MEN} and \textit{VisSim}/\textit{SemSim} test sets which average 4.2 and 4.8 respectively. Therefore, with the exceptions of \textit{MEN}, \textit{VisSim} and \textit{SemSim}, the inclusion of multimodal information in the ZS regions is arguably less beneficial than in the VIS regions, given that visual information can only sensibly enrich representations of words that are to some extent visual.

\begin{table*}[ht]
	\footnotesize
	\centering
	\caption{Spearman correlations between model predictions and human ratings. For each test set, ALL is the whole set of word pairs, VIS are those pairs with both visual representations available, and ZS denotes its complement, i.e., zero-shot words. Boldface indicates best results per column and \# inst. the number of word pairs in ALL, VIS or ZS. It must be noted that the VIS region of the compared methods is only approximated, as they do not report the exact evaluated instances.}
	\begin{adjustbox}{max width=\textwidth}
\begin{tabular}{lcccccccccccc}
	\multicolumn{1}{l|}{} & \multicolumn{3}{c|}{Wordsim353} & \multicolumn{3}{c|}{MEN} & \multicolumn{3}{c|}{SemSim} & \multicolumn{3}{c|}{VisSim} \\ \cline{2-13} 
	\multicolumn{1}{l|}{} & ALL & VIS & \multicolumn{1}{c|}{ZS} & ALL & VIS & \multicolumn{1}{c|}{ZS} & ALL & VIS & \multicolumn{1}{c|}{ZS} & ALL & VIS & \multicolumn{1}{c|}{ZS} \\ \hline
	\multicolumn{1}{l|}{Silberer \& Lapata 2014} &  &  & \multicolumn{1}{c|}{} &  & 0.7 & \multicolumn{1}{c|}{} &  & 0.64 & \multicolumn{1}{c|}{} &  &  & \multicolumn{1}{c|}{} \\
	\multicolumn{1}{l|}{Lazaridou et al. 2015} &  &  & \multicolumn{1}{c|}{} & 0.75 &0.76  & \multicolumn{1}{c|}{} & 0.72 & 0.72 & \multicolumn{1}{c|}{} & 0.63 & 0.63 & \multicolumn{1}{c|}{} \\
	\multicolumn{1}{l|}{Kiela \& Bottou 2014} &  & 0.61 & \multicolumn{1}{c|}{} &  & 0.72 & \multicolumn{1}{c|}{} &  &  & \multicolumn{1}{c|}{} &  &  & \multicolumn{1}{c|}{} \\ \hline
	\multicolumn{1}{l|}{GloVe} & \textbf{0.712} & 0.632 & \multicolumn{1}{c|}{\textbf{0.705}} & 0.805 & 0.801 & \multicolumn{1}{c|}{0.801} & 0.753 & 0.768 & \multicolumn{1}{c|}{0.701} & 0.591 & 0.606 & \multicolumn{1}{c|}{0.54} \\
	\multicolumn{1}{l|}{\textit{CNN}$_{avg}$} & - & 0.448 & \multicolumn{1}{c|}{-} & - & 0.593 & \multicolumn{1}{c|}{-} & - & 0.534 & \multicolumn{1}{c|}{-} & - & 0.56 & \multicolumn{1}{c|}{-} \\
	\multicolumn{1}{l|}{\textit{CONC}} & - & 0.606 & \multicolumn{1}{c|}{-} & - & 0.8 & \multicolumn{1}{c|}{-} & - & 0.734 & \multicolumn{1}{c|}{-} & - & 0.651 & \multicolumn{1}{c|}{-} \\ \hline
	\multicolumn{1}{l|}{\textit{MAP}$_{NN}$} & \multicolumn{1}{l}{0.443} & \multicolumn{1}{l}{0.534} & \multicolumn{1}{l|}{0.391} & \multicolumn{1}{l}{0.703} & \multicolumn{1}{l}{0.761} & \multicolumn{1}{l|}{0.68} & \multicolumn{1}{l}{0.729} & \multicolumn{1}{l}{0.732} & \multicolumn{1}{l|}{0.718} & \multicolumn{1}{l}{\textbf{0.658}} & \multicolumn{1}{l}{\textbf{0.659}} & \multicolumn{1}{l|}{\textbf{0.655}} \\
	\multicolumn{1}{l|}{\textit{MAP}$_{lin}$} & \multicolumn{1}{l}{0.402} & \multicolumn{1}{l}{0.539} & \multicolumn{1}{l|}{0.366} & \multicolumn{1}{l}{0.701} & \multicolumn{1}{l}{0.774} & \multicolumn{1}{l|}{0.674} & \multicolumn{1}{l}{0.738} & \multicolumn{1}{l}{0.738} & \multicolumn{1}{l|}{0.74} & \multicolumn{1}{l}{0.646} & \multicolumn{1}{l}{0.644} & \multicolumn{1}{l|}{0.651} \\
	\multicolumn{1}{l|}{\textit{MAP-C}$_{NN}$} & 0.687 & 0.644 & \multicolumn{1}{c|}{0.673} & \textbf{0.813} & \textbf{0.82} & \multicolumn{1}{c|}{\textbf{0.806}} & 0.783 & \textbf{0.791} & \multicolumn{1}{c|}{0.754} & 0.65 & 0.657 & \multicolumn{1}{c|}{0.626} \\
	\multicolumn{1}{l|}{\textit{MAP-C}$_{lin}$} & 0.694 & \textbf{0.649} & \multicolumn{1}{c|}{0.684} & 0.811 & 0.819 & \multicolumn{1}{c|}{0.802} & \textbf{0.785} & \textbf{0.791} & \multicolumn{1}{c|}{\textbf{0.764}} & 0.641 & 0.647 & \multicolumn{1}{c|}{0.623} \\ \hline
	\multicolumn{1}{l|}{\# inst.} & 353 & 63 & \multicolumn{1}{c|}{290} & 3000 & 795 & \multicolumn{1}{c|}{2205} & 6933 & 5238 & \multicolumn{1}{c|}{1695} & 6933 & 5238 & \multicolumn{1}{c|}{1695} \\
	\multicolumn{13}{l}{} \\
	\multicolumn{1}{l|}{} & \multicolumn{3}{c|}{Simlex999} & \multicolumn{3}{c|}{Wordsim353-rel} & \multicolumn{3}{c|}{Wordsim353-sim} & \multicolumn{3}{c|}{SimVerb-3500} \\ \cline{2-13} 
	\multicolumn{1}{l|}{} & ALL & VIS & \multicolumn{1}{c|}{ZS} & ALL & VIS & \multicolumn{1}{c|}{ZS} & ALL & VIS & \multicolumn{1}{c|}{ZS} & ALL & VIS & \multicolumn{1}{c|}{ZS} \\ \hline
	\multicolumn{1}{l|}{Silberer \& Lapata 2014} &  &  & \multicolumn{1}{c|}{} &  &  & \multicolumn{1}{c|}{} &  &  & \multicolumn{1}{c|}{} &  &  & \multicolumn{1}{c|}{} \\
	\multicolumn{1}{l|}{Lazaridou et al. 2015} & 0.4 & \textbf{0.53} & \multicolumn{1}{c|}{} &  &  & \multicolumn{1}{c|}{} &  &  & \multicolumn{1}{c|}{} &  &  & \multicolumn{1}{c|}{} \\
	\multicolumn{1}{l|}{Kiela \& Bottou 2014} &  &  & \multicolumn{1}{c|}{} &  &  & \multicolumn{1}{c|}{} &  &  & \multicolumn{1}{c|}{} &  &  & \multicolumn{1}{c|}{} \\ \hline
	\multicolumn{1}{l|}{GloVe} & 0.408 & 0.371 & \multicolumn{1}{c|}{\textbf{0.429}} & \textbf{0.644} & 0.759 & \multicolumn{1}{c|}{\textbf{0.619}} & \textbf{0.802} & 0.688 & \multicolumn{1}{c|}{\textbf{0.783}} & 0.283 & 0.32 & \multicolumn{1}{c|}{0.282} \\
	\multicolumn{1}{l|}{\textit{CNN}$_{avg}$} & - & 0.406 & \multicolumn{1}{c|}{-} & - & 0.422 & \multicolumn{1}{c|}{-} & - & 0.526 & \multicolumn{1}{c|}{-} & - & 0.235 & \multicolumn{1}{c|}{-} \\
	\multicolumn{1}{l|}{\textit{CONC}} & - & 0.442 & \multicolumn{1}{c|}{-} & - & 0.665 & \multicolumn{1}{c|}{-} & - & 0.664 & \multicolumn{1}{c|}{-} & - & 0.437 & \multicolumn{1}{c|}{-} \\ \hline
	\multicolumn{1}{l|}{\textit{MAP}$_{NN}$} & \multicolumn{1}{l}{0.322} & \multicolumn{1}{l}{0.451} & \multicolumn{1}{l|}{0.296} & \multicolumn{1}{l}{0.33} & \multicolumn{1}{l}{0.606} & \multicolumn{1}{l|}{0.267} & \multicolumn{1}{l}{0.536} & \multicolumn{1}{l}{0.599} & \multicolumn{1}{l|}{0.475} & \multicolumn{1}{l}{0.213} & \multicolumn{1}{l}{\textbf{0.513}} & \multicolumn{1}{l|}{0.21} \\
	\multicolumn{1}{l|}{\textit{MAP}$_{lin}$} & \multicolumn{1}{l}{0.322} & \multicolumn{1}{l}{0.412} & \multicolumn{1}{l|}{0.286} & \multicolumn{1}{l}{0.28} & \multicolumn{1}{l}{0.553} & \multicolumn{1}{l|}{0.243} & \multicolumn{1}{l}{0.505} & \multicolumn{1}{l}{0.569} & \multicolumn{1}{l|}{0.477} & \multicolumn{1}{l}{0.212} & \multicolumn{1}{l}{0.338} & \multicolumn{1}{l|}{0.21} \\
	\multicolumn{1}{l|}{\textit{MAP-C}$_{NN}$} & 0.405 & 0.404 & \multicolumn{1}{c|}{0.417} & 0.623 & 0.778 & \multicolumn{1}{c|}{0.589} & 0.769 & 0.696 & \multicolumn{1}{c|}{0.745} & \textbf{0.286} & 0.49 & \multicolumn{1}{c|}{0.284} \\
	\multicolumn{1}{l|}{\textit{MAP-C}$_{lin}$} & \textbf{0.41} & 0.388 & \multicolumn{1}{c|}{0.422} & 0.629 & \textbf{0.797} & \multicolumn{1}{c|}{0.601} & 0.781 & \textbf{0.698} & \multicolumn{1}{c|}{0.766} & \textbf{0.286} & 0.371 & \multicolumn{1}{c|}{\textbf{0.285}} \\ \hline
	\multicolumn{1}{l|}{\# inst.} & 999 & 261 & \multicolumn{1}{c|}{738} & 252 & 28 & \multicolumn{1}{c|}{224} & 203 & 45 & \multicolumn{1}{c|}{158} & 3500 & 41 & \multicolumn{1}{c|}{3459}
\end{tabular}
	\end{adjustbox}
	\label{tab:results}
\end{table*}

\begin{figure}
	\begin{center}
		\includegraphics[scale=0.1]{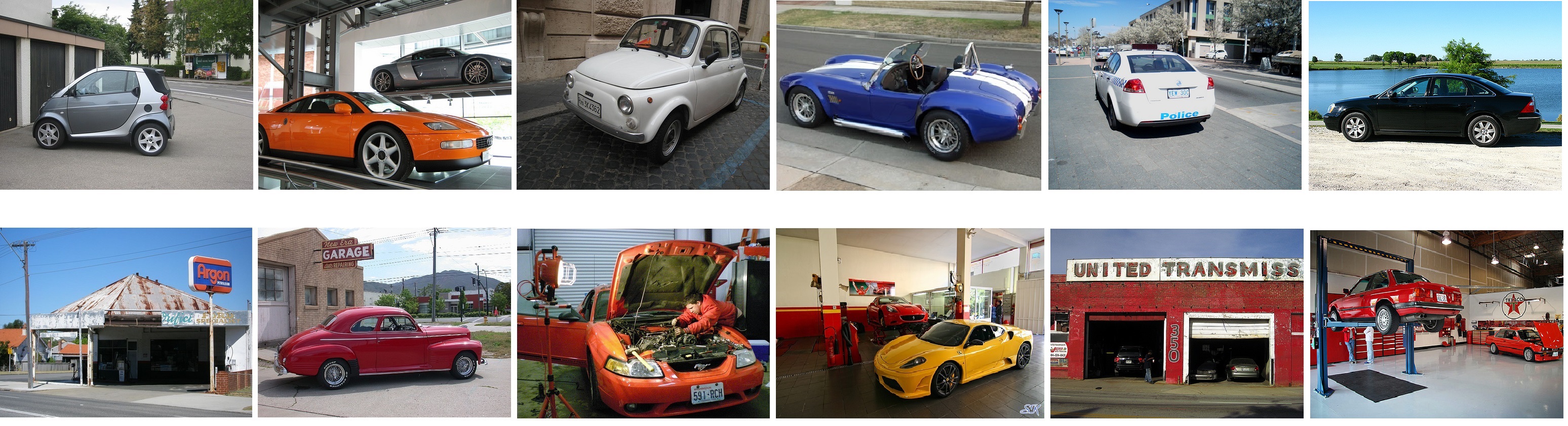}
	\end{center}
	\caption{Sample images from ``car" (top row) and ``garage" (bottom row) synsets of ImageNet.}
	\label{car-garage}
\end{figure}

Both \textit{MAP}$_{NN}$ and \textit{MAP}$_{lin}$ exhibit an overall gain in \textit{MEN} and in the VIS region of \textit{Wordsim353-rel}. It might seem counter-intuitive that vision can help to improve \textbf{\textit{relatedness}} understanding. However, a closer look reveals that visual features generally account for object co-occurrences, which is often a good indicator of their relatedness (e.g., between ``car" and ``garage" in Fig. \ref{car-garage}). For instance, in \textit{MEN}, the human relatedness rating between ``car" and ``garage" is 8.2 while GloVe's score is only 5.4. However, \textit{CNN}$_{avg}$'s rating is 8.7 and that of \textit{MAP}$_{lin}$ is 8.4---closer to the human score. 

Crucially, \textit{MAP-C}$_{NN}$ and \textit{MAP-C}$_{lin}$ significantly improve the performance of GloVe in all seven \textit{\textbf{VIS regions}} (p $\approx$ 0.008), with an average improvement of 4.6$\%$ for \textit{MAP-C}$_{NN}$. Conversely, the concatenation of GloVe with the original visual vectors (\textit{CONC}) does not improve GloVe (p $\approx$ 0.7)---worsening it in 4 out of 7 test sets---suggesting that simple concatenation without seeking the association between modalities might be suboptimal. Moreover, the concatenation of the mapped visual vectors with GloVe (\textit{MAP-C}$_{NN}$) outperforms the concatenation of the original visual vectors with GloVe (\textit{CONC}) in 6 out of 7 test sets (p $\approx$ 0.06), which supports our claim that the mapped visual vectors are semantically richer than the original visual vectors.

\section{Conclusions} 
We have presented a cognitively-inspired method capable of generating multimodal representations in a fast and simple way. In a variety of similarity tasks and seven benchmark tests, our method generally outperforms unimodal baselines and state-of-the-art multimodal methods. Moreover, the performance gain in zero-shot settings indicates that the method generalizes well and learns relevant cross-modal associations. Finally, the overall performance supports the claim that our approach builds more ``human-like" concept representations. Ultimately, the present work sheds light on fundamental questions of natural language understanding such as whether the nature of the knowledge representation obtained by the fusion of vision and language should be static and additive (e.g., concatenation without associating modalities) or rather re-constructive and associative.

\bibliographystyle{ieeetr}
\bibliography{NIPS_workshop2016}

\end{document}